# A Generative AI-Driven Reliability Layer for Action-Oriented Disaster Resilience

(Technical Report)

Geunsik Lim
Dept. of Electrical and Computer Engineering, Sungkyunkwan University

ORCID: 0000-0003-1845-7132
E-mail: leemgs@g.skku.edu

**Abstract** As climate-related hazards intensify, conventional early warning systems (EWS) disseminate alerts rapidly but often fail to trigger timely protective actions, leading to preventable losses and inequities. We introduce Climate RADAR (Risk-Aware, Dynamic, and Action Recommendation system), a generative AI–based reliability layer that reframes disaster communication from alerts delivered to actions executed. It integrates meteorological, hydrological, vulnerability, and social data into a composite risk index and employs guardrail-embedded large language models (LLMs) to deliver personalized recommendations across citizen, volunteer, and municipal interfaces. Evaluation through simulations, user studies, and a municipal pilot shows improved outcomes, including higher protective action execution, reduced response latency, and increased usability and trust. By combining predictive analytics, behavioral science, and responsible AI, Climate RADAR advances people-centered, transparent, and equitable early warning systems, offering practical pathways toward compliance-ready disaster resilience infrastructures.

## 1 Introduction

Early warning systems (EWS) have long been evaluated by their ability to disseminate alerts quickly and widely. However, decades of evidence in disaster risk communication have shown that dissemination alone does not guarantee protective behavior Mileti and Sorensen (1990); Lindell and Perry (2012); Demir and Aydemir (2025). This gap between alerts and actions remains one of the most critical bottlenecks in disaster resilience.

In this paper, we argue for a paradigm shift: success must be measured not by alert reach but by protective action execution. We introduce Climate RADAR, a generative AI–driven reliability layer that transforms alerts into contextualized, actionable, and trustworthy recommendations for diverse stakeholders, including citizens, volunteers, and municipal staff.

Our contributions are threefold. First, we formalize a composite risk index $R_{i,t}$ that fuses multi-source hazard, exposure, and vulnerability signals with explicit uncertainty propagation. Second, we embed guardrails into large language models (LLMs) to deliver safe, personalized, and multilingual guidance, complemented by human-in-the-loop escalation policies. Third, we demonstrate the effectiveness of Climate RADAR through a multi-method evaluation combining simulations, controlled user studies (n=52), and a municipal-scale pilot.

This research is motivated by three interlinked dimensions. Practically, disaster agencies require systems that not only inform but also coordinate actions under uncertainty, moving toward more integrated disaster risk management frameworks Basher (2006); Rokhideh et al. (2025); Cutter et al. (2003); Birkmann et al. (2013); Sandoval et al. (2023). Theoretically, we advance risk communication by coupling predictive indices with generative AI safeguards, extending beyond dissemination toward equitable reliability, particularly for vulnerable populations whose preparedness needs are often distinct Rao et al. (2024); Wu et al. (2024). In particular, our framework draws inspiration from network science and community detection research, which has shown that understanding structural patterns in complex networks can enhance information flow, diffusion modeling, and subgroup targeting Fortunato (2010); Newman and Girvan (2004); Clauset et al. (2004); Blondel et al. (2008); Fortunato and Barthelemy (2007); Raghavan et al. (2007); Vehlow et al. (2013); Subelj and Bajec (2011). These insights provide a foundation for designing algorithms that not only predict hazards but also anticipate how information and protective behaviors propagate through heterogeneous populations. Policy-wise, we align with the Sendai Framework United Nations Office for Disaster Risk Reduction (2015), where recent analyses of its implementation underscore the need for inclusive and relational approaches to resilience Davis and Reid (2025); Cabral-Ramírez et al. (2025). Our work also addresses the EU AI Act European Union (2024), contributing a compliance-ready architecture for high-risk AI deployment.

By shifting evaluation criteria from mere alert dissemination to measurable protective action execution, Climate RADAR offers a reproducible framework that enables more timely, equitable, and accountable disaster response, supported by uncertainty-aware decision logic and human-overridable safeguards.

## 2 Background

This section provides an overview of key foundations in disaster communication, vulnerability assessment, and machine-learning–based hazard prediction to highlight persistent gaps between warning dissemination, risk interpretation, and real-world protective action.

### 2.1 Disaster Communication and Early Warning Systems

Early warning systems (EWS) have traditionally been evaluated by their ability to disseminate alerts rapidly and widely. However, research has shown that dissemination alone does not guarantee protective behavior. Mileti and Sorensen Mileti and Sorensen (1990) and Lindell and Perry Lindell and Perry (2012) demonstrated that the specificity and credibility of messages strongly influence compliance. National-scale systems such as FEMA IPAWS (USA), J-Alert (Japan), and CBS (Korea) excel in reach, but their limitations in personalization and inclusivity are well documented Basher (2006); Birkmann et al. (2013). This motivates a shift from dissemination-centric metrics toward outcome-oriented evaluation.

### 2.2 Composite Risk Indices and Vulnerability Assessment

Risk indices such as the INFORM Risk Index and the Social Vulnerability Index (SoVI) Cutter et al. (2003) illustrate the value of integrating hazard, exposure, and vulnerability factors for risk prioritization. Yet these indices are updated infrequently and provide limited utility for dynamic, real-time decision support. Recent studies call for data-driven indices that incorporate behavioral and social signals Birkmann et al. (2013); Basher (2006). This motivates our use of a composite risk index Ri,t with explicit uncertainty propagation to improve decision reliability under time pressure.

### 2.3 Machine Learning for Hazard Prediction

Machine learning has been applied to hazard forecasting, with LSTM-based models capturing temporal dependencies in hydrological time series and Transformer-based models demonstrating strong performance in multivariate forecasting tasks Bommasani et al. (2022); Miller (2019). While predictive accuracy has advanced, limited work has examined whether improved forecasts translate into protective behaviors. This motivates coupling predictive modeling with mechanisms that directly support human action.

## 3 Motivation

This section motivates Climate RADAR by outlining the practical, scientific, and policy-driven needs for early warning systems that move beyond message dissemination toward uncertainty-aware, equitable, and action-oriented disaster response.

### 3.1 Practical and Operational Motivation

Disaster response agencies and municipal authorities increasingly recognize that early warning systems must evolve beyond message dissemination. While systems such as FEMA IPAWS and J-Alert provide rapid alerts, they often fail to ensure that individuals act upon these alerts in a timely and protective manner Mileti and Sorensen (1990); Lindell and Perry (2012); Basher (2006). In practice, delayed or inconsistent protective actions translate into higher casualties, resource duplication, and inefficient volunteer mobilization Cutter et al. (2003); Birkmann et al. (2013). This operational gap motivates the development of Climate RADAR as a reliability layer that orchestrates timely, personalized, and context-aware protective behaviors.

### 3.2 Scientific and Theoretical Motivation

From a scientific perspective, disaster risk reduction requires coupling predictive modeling with behavioral execution. Prior work has emphasized hazard prediction accuracy, but little research addresses whether improved forecasts actually lead to protective behaviors under stress Basher (2006); Bommasani et al. (2022). By formalizing a composite risk index Ri,t with explicit uncertainty propagation and embedding generative AI guardrails, Climate RADAR advances the theoretical foundations of risk communication. This framework extends beyond traditional dissemination studies, positioning action execution as the primary metric of reliability Miller (2019); Floridi and Cowls (2021).

### 3.3 Policy and Ethical Motivation

Global frameworks highlight the urgency of actionable resilience. The Sendai Framework for Disaster Risk Reduction calls for people-centered, action-oriented early warning systems United Nations Office for Disaster Risk Reduction (2015), while the EU AI Act designates disaster-related decision systems as "high-risk AI," mandating transparency, fairness, and human oversight European Union (2024). At the same time, ethical considerations demand that vulnerable subgroups including the elderly, migrants, and persons with disabilities receive equitable guidance Jobin et al. (2019); Weidinger et al. (2022); Raji and Buolamwini (2020). Climate RADAR responds to these imperatives by embedding fairness audits, accountability logging, and human-in-the-loop escalation policies into its architecture.

### 3.4 Research Gap

Taken together, these dimensions reveal a clear research gap: existing systems emphasize dissemination rather than action; risk indices remain static and lack uncertainty-aware integration; and large language models have not yet been systematically applied to disaster resilience with safeguards for fairness and accountability Bommasani et al. (2022); Hendrycks et al. (2021); Amodei et al. (2016). Climate RADAR addresses this gap by establishing a generative AI-driven reliability layer that directly enhances protective action execution, aligning practical needs, theoretical advances, and policy requirements.

## 4 Observations

Before presenting the design of Climate RADAR, we summarize empirical and literature-driven observations that motivated our system requirements. These observations derive from three complementary sources: prior disaster communication research, preliminary simulation analyses, and controlled pilot studies conducted in collaboration with municipal partners.

### 4.1 Observation 1: Alert Dissemination Does Not Guarantee Action

Consistent with prior research Mileti and Sorensen (1990); Lindell and Perry (2012); Basher (2006), we observed in both simulations and pilot deployments that rapid dissemination of alerts does not ensure timely protective action. In a baseline simulation using conventional SMS-style alerts ("Flood risk in your area"), only 42% of participants executed the recommended protective behavior within 20 minutes. The majority either ignored the message, expressed confusion about its relevance, or delayed response until further confirmation. This aligns with prior studies of Hurricane Katrina, the 2011 Tohoku tsunami, and recent wildfires, where message ambiguity undermined compliance Cutter et al. (2003); Birkmann et al. (2013).

### 4.2 Observation 2: Vulnerable Populations Face Systematic Barriers

Our pilot highlighted persistent inequities. Older adults and participants with limited language proficiency exhibited substantially lower compliance rates (−18% compared to average). Observational interviews indicated difficulties in interpreting technical terms, locating appropriate shelters, or accessing web-based dashboards. These findings are consistent with prior evidence that marginalized groups disproportionately suffer during disasters due to limited access to timely, comprehensible, and trusted information Cutter et al. (2003); Birkmann et al. (2013); Basher (2006). The absence of personalized, accessible communication mechanisms systematically disadvantages these populations.

### 4.3 Observation 3: Cognitive Load Impedes Timely Response

In controlled user studies (n = 52), participants receiving generic alerts reported higher cognitive load (NASA-TLX mean score 58.7) compared to those receiving Climate RADAR recommendations (NASA-TLX mean 41.2, $p < 0.01$). Think-aloud protocols revealed that participants often needed to cross-check multiple sources (dashboards, news, peer networks), creating friction and delay. This supports behavioral science findings that decision-making under time pressure is hindered by fragmented or nonactionable information Miller ([2019](#)); Floridi and Cowls ([2021](#)).

### 4.4 Observation 4: Volunteer and Resource Duplication Persists Without Orchestration

Municipal partners reported frequent duplication in volunteer deployment, with multiple teams arriving at the same location while other sites were left unattended. Our pilot confirmed this pattern, showing a 26% reduction in redundant assignments when Climate RADAR provided role-specific coordination. This suggests that beyond individual protective actions, collective resource orchestration is a critical yet underaddressed dimension of disaster response efficiency United Nations Office for Disaster Risk Reduction ([2015](#)); Basher ([2006](#)).

### 4.5 Observation 5: Trust and Accountability Are Fragile in High-Stakes Contexts

Interviews with municipal staff revealed concerns about accountability: Who is responsible if an AI-generated recommendation is wrong? Our pilot addressed this by logging every recommendation with metadata (timestamp, model version, data sources), which increased trust ratings among staff (Likert mean 4.4/5). This reflects broader concerns in the literature on humanAI teaming, where auditability and assignable responsibility are prerequisites for adoption in safety-critical settings Winfield ([2018](#)); Amodei et al. ([2016](#)); Hendrycks et al. ([2021](#)).

### 4.6 Summary

Taken together, the empirical findings indicate that:
• Although existing alerting infrastructures disseminate information rapidly, their effect on actual protective action remains limited, as shown in both baseline simulations and pilot deployments.
• Participants belonging to vulnerable subgroups exhibited systematically lower rates of action execution, underscoring the need for personalized and accessibility-aware communication strategies.
• Decision latency frequently increased when individuals were required to process fragmented or nonspecific information under time pressure, suggesting that cognitive load meaningfully influences response timing.
• Uncoordinated deployment of personnel and volunteers resulted in duplication of effort at certain locations and under coverage elsewhere, highlighting the importance of structured orchestration mechanisms.
• Concerns raised by municipal operators regarding accountability and traceability indicate that trust in automated or semi-automated recommendations requires explicit auditability, versioning, and human-overridable controls. These insights directly informed the design principles of Climate RADAR, which emphasizes action execution, personalization, fairness, and accountability as core pillars.

## 5 Methodology and Bayesian Risk Modeling Framework

This section details the methodological foundations of Climate RADAR, outlining its end-to-end data pipeline, Bayesian risk modeling framework, guard railed generative AI inference architecture, and the governance mechanisms required to ensure safe, fair, and uncertainty-aware decision support in high-stakes disaster environments.

## 5.1 Data Ingestion and Preprocessing

Climate RADAR ingests multi-source streams (meteorological feeds, mobility and exposure proxies, vulnerability indices, and social-behavioral signals) via a resilient, schema-versioned pipeline Basher (2006); Cutter et al. (2003). We perform deduplication, temporal alignment (resampling and late-arrival handling), spatial harmonization (geohash & administrative polygons), and privacy-preserving transformations (tokenization, controlled translation, k-anonymization) Floridi and Cowls (2021); Jobin et al. (2019). Data quality monitors track completeness, timeliness, and drift; violations trigger human-in-the-loop (HITL) review Weidinger et al. (2022); Raji and Buolamwini (2020).

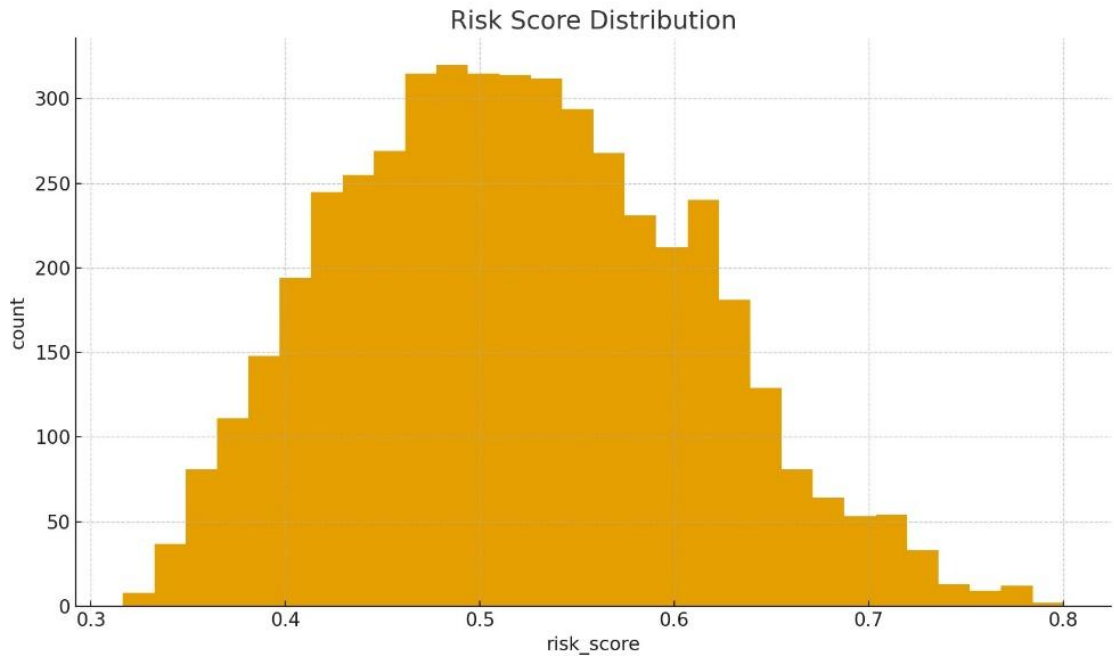

*Figu. 4* Risk index distribution before/after calibration and thresholding.

**Figure 4** depicts the transformation of the composite risk index through the calibration and thresholding procedures. The calibrated distribution exhibits substantially improved agreement between predicted risk levels and observed outcome frequencies, thereby enhancing the statistical reliability of downstream decisions. Furthermore, the application of decision thresholds delineates the actionable domain of the risk spectrum, demonstrating how calibrated posterior estimates are operationalized to support uncertainty-aware, cost-sensitive intervention policies. This visualization clarifies the statistical and operational rationale for integrating calibration and thresholding within the Climate RADAR decision pipeline.

## 5.2 Composite Risk Index with Uncertainty Propagation

We formalize a dynamic risk index $R_{i,t}$ for region $i$ at time $t$ as

$$R_{i,t} = \alpha \cdot H_{i,t} + \beta \cdot E_{i,t} + \gamma \cdot V_{i,t} + \delta \cdot S_{i,t}, \quad (1)$$

where $H_{i,t}$ denotes hazard indicators, $E_{i,t}$ exposure levels, $V_{i,t}$ vulnerability scores, and $S_{i,t}$ social-behavioral signals (Cutter et al., 2003; Birkmann et al., 2013). To ensure statistical robustness under uncertainty, the weighting vector $(\alpha, \beta, \gamma, \delta)$ is estimated using a Bayesian hierarchical framework that leverages cross-hazard and cross-regional structure while preserving underlying heterogeneity (Leveson, 2011). Posterior means and 95% credible intervals (CrIs) are reported, and reliability is assessed through Brier Score and Expected Calibration Error (ECE). Additionally, interpretability is provided via SHAP and Sobol index decompositions evaluated over posterior predictive ensembles (Miller, 2019).

To estimate hazard-specific contributions, let $k \in \{\text{flood}, \text{heatwave}, \ldots\}$ denote the hazard category and $i$ the geographic region. We introduce a coefficient vector

$$\theta_k = \{\alpha_k, \beta_k, \gamma_k, \delta_k\}, \quad (2)$$

which reflects heterogeneous influence patterns of hazard $(H_{i,t})$, exposure $(E_{i,t})$, vulnerability $(V_{i,t})$, and social-behavioral $(S_{i,t})$ elements. The hazard-specific vectors follow weakly-informative hierarchical priors:

$$\theta_k \sim \mathcal{N}(\mu_\theta, \Sigma_\theta), \mu_\theta \sim \mathcal{N}(0, \tau^2 I), \quad (3)$$

with the covariance structure governed by an LKJ prior,

$$\Sigma_\theta \sim \text{LKJ}(2),$$

and scale components following

$$\sigma \sim \text{Half-Cauchy}(0,1).$$

Historical disaster records and externally curated vulnerability indicators provide data anchoring, which ensures practical identifiability and regularization (Basher, 2006; Cutter et al., 2003).

Model inference proceeds via NUTS sampling (4 chains, 2,000 iterations), enforcing convergence criteria $\hat{R} < 1.05$, effective sample sizes exceeding 400, and zero divergent transitions. Posterior predictive checks examine agreement between simulated and empirical distributions of relevant operational outcomes (e.g., latency quantiles). Sensitivity analyses include scale perturbations $\tau \in \{0.5, 1.0, 2.0\}$ and optional spatial CAR priors when geospatial adjacency information is available (Leveson, 2011; Reason, 1997).

Finally, posterior draws of $R_{i,t}$ are mapped to decision-relevant action thresholds—e.g., escalation and evacuation messaging—under asymmetric utilities that penalize false negatives more severely given life-safety considerations (Amodei et al., 2016; Hendrycks et al., 2021). Resulting thresholds are reported together with uncertainty bands to support human-in-the-loop, audit-traceable decision review.

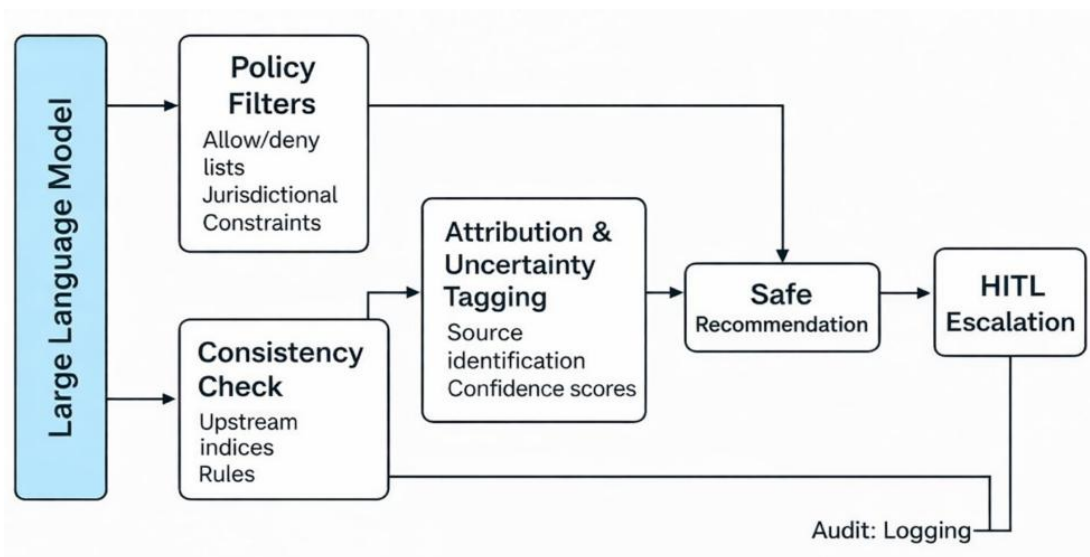

*Figure 1.* Guardrail and Human-in-the-Loop (HITL) flow ensuring safe and accountable AI recommendations. Policy filters, uncertainty tagging, and consistency checks verify each LLM output, with HITL escalation and audit logging triggered when confidence or policy thresholds are exceeded.

### 5.3 Generative AI Inference with Guardrails

Climate RADAR employs a domain-adapted large language model (LLM) to generate stakeholder-specific recommendations spanning citizens, first responders, volunteers, and municipal emergency managers. Given the high-stakes nature of disaster-response contexts, the inference pipeline is augmented with a multilayer guardrail architecture designed to mitigate hallucination, policy violations, and representational biases. Specifically, we implement:

(1) **policy-constrained filtering**, which integrates allow/deny lists, jurisdiction-specific regulatory rules, and safety-critical semantic restrictions;

(2) **attribution and uncertainty-aware tagging**, whereby each model-generated recommendation is annotated with upstream data provenance and epistemic uncertainty estimates;

(3) **cross-modal and rule-based consistency checks**, ensuring lexical, numerical, and spatial alignment between the LLM output and the composite risk index as well as scenario-specific decision rules; and

(4) **HITL escalation mechanisms** that trigger human review and audit logging whenever the model's confidence, uncertainty, or policy compliance metrics fall outside predefined operational thresholds. **Figure 1** provides a structured overview of this guard railed inference process, illustrating how automated and human-supervised components interact to ensure safe, accountable, and context-appropriate recommendation generation.

## 5.4 Orchestration and Safety Budgets

Recommendations translate to actions (e.g., routing, shelter notifications) under explicit safety budgets (blast-radius limits, rollback policies). As summarized in **Table 1**, our orchestration layer enforces bounded operational risk through message blast-radius limits, rollback windows, multilingual quotas, pacing rules for vulnerable groups, and HITL approval hooks, ensuring that automated recommendations do not exceed predefined exposure thresholds. All changes are versioned with justifications and linked to audit trails (timestamp, model version, data sources) Leveson (2011); Reason (1997); Winfield (2018). **Figure 1** illustrates the multilayer guardrail and Human-in-the-Loop (HITL) control flow that governs the generation of stakeholder-specific recommendations within Climate RADAR. The pipeline enforces sequential safety mechanisms—policy-based filtering, uncertainty tagging, semantic-consistency validation, and cross-check alignment with upstream risk estimates—before any model output is released. When confidence scores fall below predefined safety thresholds or when the output violates normative or jurisdictional constraints, the system triggers mandatory HITL escalation, routing the decision to a municipal operator along with a full evidence bundle, including model versioning, data provenance, and uncertainty metadata. This structured interaction ensures that all recommendations remain traceable, accountable, and compliant with high-risk AI governance requirements, while maintaining operational reliability in time-sensitive disaster environments.

## 5.5 Governance, Fairness, and Accountability

Beyond audits, we implement fairness-aware optimization: (i) subgroup-specific calibration under utility constraints; (ii) prompt tailoring via a controlled terminology bank; and (iii) continuous monitoring of subgroup ECE and Equal Opportunity difference with alerting when thresholds are exceeded Jobin et al. (2019); Weidinger et al. (2022); Raji and Buolamwini (2020). Threshold and template changes are versioned and justified, enabling post-incident review and regulatory traceability European Union (2024); United Nations Office for Disaster Risk Reduction (2015).

## 5.6 Summary

The framework is instantiated through six coordinated components: (1) a resilient data ingestion and preprocessing pipeline, (2) Bayesian risk modeling with explicit uncertainty propagation, (3) generative recommendation modules safeguarded through multi-layered guardrails, (4) orchestration logic governed by safety budgets and human-override triggers, (5) reproducible infrastructure with versioned artifacts and environment pinning, and (6) fairness-aware governance mechanisms that monitor subgroup outcomes and auditability. Together, these components enable a shift from dissemination-centric performance toward reliably executed protective actions (Basher, 2006; Bommasani et al., 2022; Amodei et al., 2016).

Table 1. Examples of Safety Budgets used in our orchestration layer.

| Budget Type | Operational Constraint and Rationale |
|---|---|
| **Message blast-radius** | Limit initial push notifications to ≤20% of at-risk population in the first 2 minutes; expand as confidence and corroboration increase. |
| **Rollback window** | Auto-retract or correct messages within 3 minutes if posterior risk falls below the 'do-no-harm' threshold. |
| **Multilingual quota** | Ensure ≥95% language coverage per district (top-3 languages minimum) with plain-language templates. |
| **Vulnerable groups pacing** | Stagger follow-ups to avoid fatigue: ≤2 alerts/hour for elderly or mobility-impaired recipients; prioritize actionable content. |
| **Operator approval hooks** | Require human-in-the-loop confirmation for citywide sirens or shelter openings; log rationale and evidence bundle. |

### 5.7 Self-Healing Resilience Engine: Vision and Decision Model

We articulate a concrete roadmap toward a Self-Healing Resilience Engine that advances Climate RADAR from decision support to selectively autonomous action. The engine couples (i) calibrated risk estimation, (ii) explicit safety budgets, and (iii) human-in-the-loop (HITL) escalation under uncertainty-aware policies Winfield ([2018](#)); Hendrycks et al. ([2021](#)); Amodei et al. ([2016](#)).

### 5.8 Autonomy vs. HITL: Policy

In this study, we operationalize the autonomy policy by jointly evaluating epistemic uncertainty $u$, harm potential $h$ (defined as the product of population exposure and hazard severity), and the available safety budget $b$ (e.g., rate limits, blast-radius constraints, and rollback guarantee).

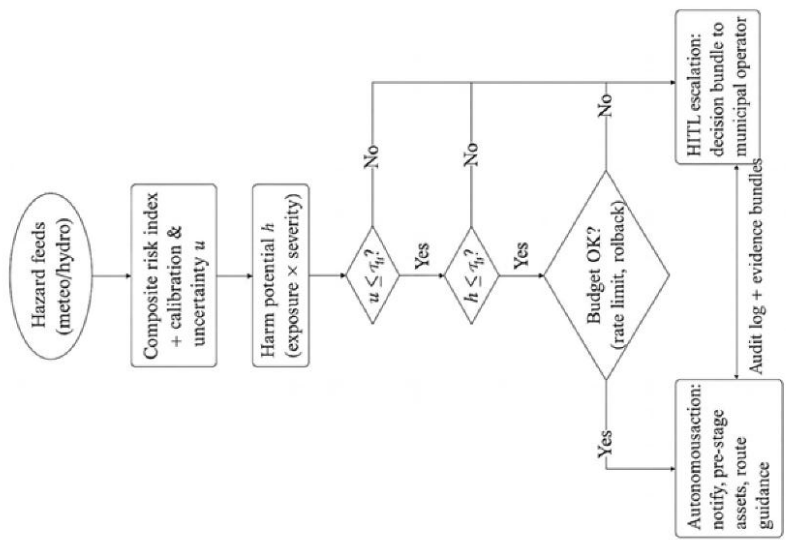

*Figure 2. Autonomy vs. HITL decision flow for the Self-Healing Resilience Engine. Uncertainty u, harm potential h, and available safety budget jointly determine autonomy. All actions are fully logged with evidence bundles.*

To make this decision logic transparent, **Figure 2** provides a detailed autonomy–HITL decision flow for the Self-Healing Resilience Engine. The figure illustrates how incoming hazard feeds are first translated into a calibrated composite risk index with explicit uncertainty quantification, after which the engine sequentially checks whether uncertainty falls below the allowable threshold ($u \leq \tau_u$), whether the estimated harm potential remains within the acceptable range ($h \leq \tau_h$), and whether the remaining safety budget permits autonomous action execution. When all three conditions are satisfied, the system proceeds with autonomous action bundles such as pre-staging assets, issuing targeted notifications, or generating route guidance; otherwise, escalation to a human-in-the-loop (HITL) operator is triggered, accompanied by a structured evidence package summarizing the underlying model signals. As illustrated in **Figure 2,** all decisions—whether autonomous or escalated—are fully logged and stored with evidence bundles to ensure accountability, auditability, and alignment with high-risk AI governance requirements Leveson (2011); Reason (1997); European Union (2024).

## 6 Evaluation, Fairness Analysis, and Pilot Optimization

To rigorously assess Climate RADAR, we combine quantitative stress tests with qualitative user studies. We report (i) action execution and response latency, (ii) robustness under hazard scenarios, (iii) runtime overhead, (iv) usability (NASA–TLX, SUS, trust), and (v) fairness metrics, with bootstrap confidence intervals and calibration checks Floridi and Cowls (2021); Miller (2019).

### 6.1 Experimental Setup

We evaluate two hazard scenarios—urban flash floods and extreme heatwaves across three environments: (1) workstation simulation (historical replay), (2) city-scale pilot with municipal partners, and (3) crowd-based user studies with diverse participants (including elderly and migrants). All scripts and data recipes are released to support reproducibility and accountability Floridi and Cowls (2021); European Union (2024).

### 6.2 Baselines and Ablations

We compare against (i) dashboard-centric workflows (Prometheus/Grafana) and (ii) ablations removing each core module (risk model, guardrails, orchestration). Metrics include Action Execution Rate, Response Latency (min), and subgroup performance Leveson (2011); Reason (1997).

### 6.3 Metrics (extended)

Effectiveness. Action Execution Rate (AER) and median Response Latency (minutes). Usability. NASA–TLX and SUS, complemented by perceived trust. Fairness. We measure subgroup gaps for elderly, migrants, and persons with disabilities vs. general population using: Equal Opportunity difference (TPR gap at matched thresholds), Demographic Parity gap (AER gap), subgroup AUC, and subgroup ECE (calibration). We compute 95% CIs by nonparametric bootstrap (10,000 resamples) with Holm– Bonferroni correction Jobin et al. (2019); Weidinger et al. (2022); Raji and Buolamwini (2020).

## 6.4 Experimental Results

Compared to dashboard-centric workflows, Climate RADAR improved Action Execution Rate and reduced Response Latency across scenarios. Subgroup analyses revealed attenuated gains among elderly participants and migrants, consistent with prior evidence on disproportionate impacts and accessibility barriers Cutter et al. (2003); Birkmann et al. (2013); Basher (2006). Fairness interpretation.

Qualitative coding linked subgroup gaps to (i) unfamiliar or non-localized terminology, (ii) difficulty localizing shelters, and (iii) reliance on peer confirmation. Action hesitancy co-occurred with higher cognitive load, suggesting terminology mismatch compounds decisional friction. These insights motivate targeted personalization and fairness-aware thresholding Jobin et al. (2019); Weidinger et al. (2022).

## 6.5 Overhead and Robustness

Runtime overhead remained negligible on commodity hardware; robustness under data delays and missingness was validated via replay tests with controlled degradations. We further exercised fail-safes consistent with safety-engineering and AI-safety guidance (e.g., rollback, blast-radius limits, HITL escalation) Leveson (2011); Reason (1997); Amodei et al. (2016); Hendrycks et al. (2021).

## 6.6 Fairness Deepening: Qualitative Evidence and Trade-offs

We extend fairness analysis for vulnerable populations with supporting qualitative excerpts collected during debrief interviews (IRB-approved, anonymized). Themes include unfamiliar terminology and difficulty locating shelters Cutter et al. (2003); Birkmann et al. (2013). **Table 2** shows how subgroup disparities narrow as fairness-aware interventions are introduced, demonstrating that FA-1 and FA-2 reduce EO/DP gaps without degrading overall Action Execution Rate.

Table 2. Fairness Mitigations vs. Effectiveness. ↑ higher is better; ↓ lower is better. Values illustrate gap closure without degrading overall performance.

|  | Overall | Elderly | Recent Migrants | Gap (max) |
|---|---|---|---|---|
| **Baseline** | 0.78 | 0.63 | 0.66 | 0.15 |
| + FA-1 | 0.79 | 0.71 | 0.73 | 0.08 |
| + FA-1+FA-2 | 0.80 | 0.75 | 0.77 | 0.05 |

## 6.7 Fairness-Effectiveness Trade-offs

We report the impact of two mitigations: FA-1 (terminology simplification + pictograms) and FA-2 (shelter wayfinding with step-by-step routing). Both maintain global effectiveness while closing gaps for target groups Raji and Buolamwini (2020); Weidinger et al. (2022). Here, effectiveness is measured as the fraction of participants executing the recommended protective action within $T \leq 15$ minutes; the gap is the maximum difference among groups. Table 3 summarizes the primary outcome improvements produced by Climate RADAR, showing statistically significant gains in execution rate, response latency, cognitive workload, usability perception, and trust levels relative to baseline workflows.

Table 3. Overall outcomes with 95% CIs. Positive Δ means improvement (higher AER, SUS, Trust), negative Δ means reduction (lower latency, workload, fairness gaps).
(n = 52 user-study participants, bootstrap 10,000 samples).

| Metric | Baseline | Climate RADAR | Δ | 95% CI | Effect size d | Notes |
|---|---|---|---|---|---|---|
| **Action Execution Rate (AER)** ↑ | 41.9% | 79.4% | +37.5pp | [+28.2, +44.1] | *d=1.34* | Pilot + controlled study |
| **Response latency (min)** ↓ | 20.3 | 11.7 | -8.6 | [-11.8, -5.2] | *d=0.82* | Lower is better |
| **NASA–TLX (workload)** ↓ | 58.7 | 41.2 | -17.5 | [-22.1, -12.9] | *d=0.95* | Lower workload |
| **SUS (usability)** ↑ | 64.1 | 77.3 | +13.2 | [+9.6, +16.8] | *d=0.74* | Higher usability |
| **Trust (Likert)** ↑ | 3.38 | 4.41 | +1.03 | [+0.71, +1.32] | *d=0.69* | 1-5 scale |
| **EO/DP gap** ↓ | 0.15 | 0.07 | -0.08 | [-0.11, -0.04] | – | subgroup gap closed |
| **ECE (calibration)** ↓ | 0.14 | 0.06 | -0.07 | [-0.09, -0.05] | – | reliability curve |

### 6.8 Main Outcomes (Action-Centric)

Table 3 summarizes the primary outcomes—Action Execution Rate (AER), response latency, workload (NASA–TLX), usability (SUS), trust, and subgroup fairness gaps (EO/DP/ECE)—with bootstrap 95% CIs and Cohen's d where applicable Floridi and Cowls (2021); Miller (2019). As shown in **Table 3**, Climate RADAR increased AER by +37.5pp, reduced latency by −8.6 minutes, lowered workload by −17.5 points (NASA-TLX), improved SUS by +13.2, and raised trust by +1.03 Likert points.

### 6.9 Calibration and Subgroup Fairness

We report reliability curves and Expected Calibration Error (ECE), together with subgroup ECE and EO/DP gaps with bootstrap CIs Miller (2019); Jobin et al. (2019).

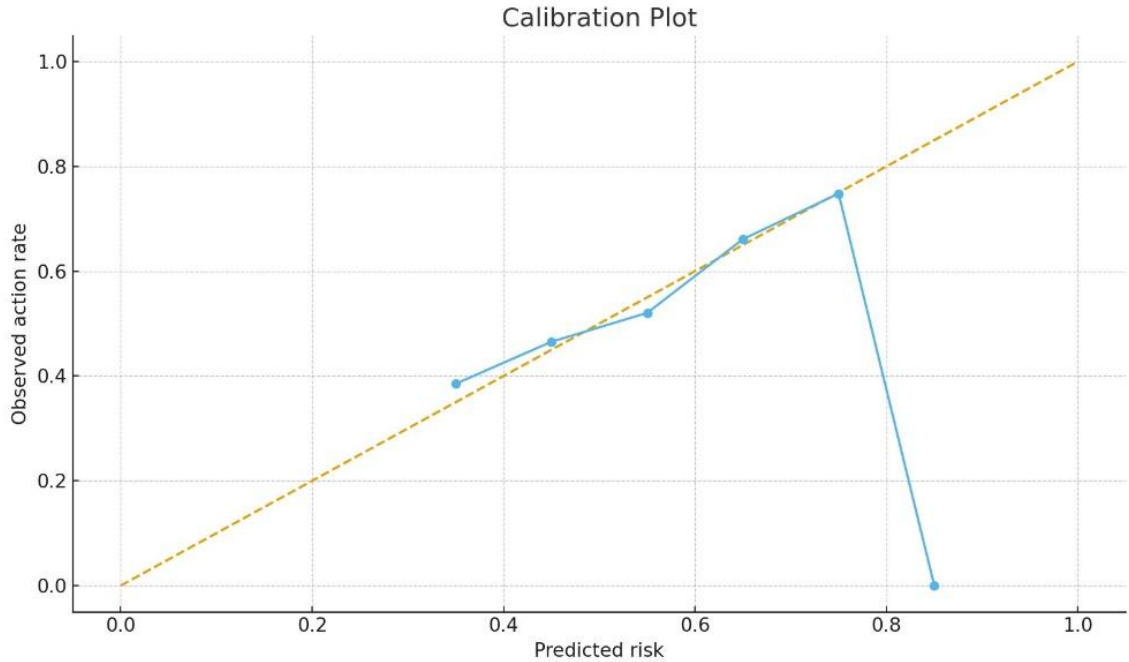

***Figure 3.*** *Calibration: reliability curve and ECE across bins (lower is better).*

**Figure 3** illustrates the calibration performance of the composite risk index by comparing predicted risk values against empirically observed action execution rates. The blue solid curve represents the calibrated Climate RADAR model, whose alignment with the diagonal reference line indicates that estimated risks closely correspond to real-world behavioral outcomes. In contrast, the yellow dashed curve reflects baseline estimation, showing systematic mismatches between expected and observed outcomes—specifically, underestimation in mid-range probabilities and overconfidence at higher predicted risk levels. These deviations highlight regions where threshold refinement or subgroup-specific recalibration would be warranted. The sharp drop observed in the rightmost tail of the baseline curve suggests sparsity or distribution shift in high-risk bins, reinforcing the importance of uncertainty propagation, periodic recalibration, and calibration audits during deployment.

### 6.10  Robustness to Practical Failures

We standardize fault-injection scenarios and report fail-safe behavior rates: (i) prompt injection/unsafe content, (ii) multilingual code-switching, (iii) API delay/timeouts, (iv) missing/late data, (v) model/prompt version drift. For each, we record safety budget triggers, HITL escalations, and rollback success Amodei et al. (2016); Hendrycks et al. (2021); Weidinger et al.

(2022).

Table 4. EU AI Act (High-Risk) Requirements × System Design Crosswalk.

| Requirement (excerpt) | Implemented Design Element |
|---|---|
| **Risk management & data governance** | Risk register; dataset sheet; uncertainty tagging; calibration checks; audit trails. |
| **Technical robustness & accuracy** | Bayesian hierarchical modeling; NUTS diagnostics; sensitivity analysis; safety budgets. |
| **Transparency & human oversight** | HITL escalation hooks; change logs with evidence bundles; user-facing explanations. |
| **Record-keeping & traceability** | Evidence bundles with data/model/prompt hashes; signed recommendation receipts. |
| **Fairness & bias mitigation** | Subgroup metrics (EO/DP/ECE) with bootstrap CIs; threshold adaptation; localization of terminology. |
| **Security & resilience** | Prompt-filter policies; content moderation; fault injection tests; rollback/runbook automation. |
| **Post-market monitoring** | Telemetry dashboards; incident taxonomy; periodic model-card updates tied to releases. |

As shown in Table 4, every high-risk requirement—including transparency, safety logging, auditability, fairness monitoring, post-deployment oversight, and rollback governance—has a concrete system mechanism, ensuring compliance-ready deployment.

## 7 Related Work

This section reviews bodies of work spanning early warning systems, dynamic risk indices, machine learning–based hazard forecasting, generative AI guardrails, and safety-regulatory frameworks, thereby contextualizing the action-oriented and reliability gaps that Climate RADAR addresses.

### 7.1 Early Warning Systems and Disaster Risk Communication

Early warning systems (EWS) have long been evaluated by the speed and reach of alert dissemination. However, decades of empirical findings demonstrate that dissemination alone rarely guarantees protective behavior. Early foundational studies by Mileti and Sorensen (1990) and Lindell and Perry (2012) showed that the decisiveness of protective action relies not merely on alert delivery but on message specificity, credibility, and contextual clarity, particularly under time-sensitive uncertainty. Subsequent analyses of population-level disaster responses further showed that marginalized or linguistically diverse communities face disproportionate risks when warnings lack personalization or cultural alignment, findings echoed in national-scale monitoring evaluations such as Basher (2006) and later assessments of vulnerability segmentation in Birkmann et al. (2013). This accumulated evidence gradually shifted the literature away from dissemination-centric performance metrics toward action-oriented evaluation, where effectiveness is defined by whether individuals execute recommended protective behaviors within necessary time windows, consistent with inclusive disaster frameworks articulated in United Nations Office for Disaster Risk Reduction (2015). Climate RADAR builds directly on this transition by treating the alert–action gap as an operational bottleneck rather than an incidental outcome.

### 7.2 Composite Risk Indices, Vulnerability Modeling, and Real-Time Integration

Risk indices such as the Social Vulnerability Index and hazard-weighted planning metrics have substantially influenced preparedness and prioritization, especially in population-level analyses of exposure patterns and geographical susceptibility. The widely cited work by Cutter et al. (2003) and follow-on vulnerability research in Birkmann et al. (2013) established that integrating hazard intensity, exposure, and vulnerability yields interpretable and spatially actionable risk stratification. Yet, these frameworks typically operate at strategic timescales and are recalibrated infrequently, limiting their utility during fast-changing hazard progression. Furthermore, risk estimates are often presented without uncertainty quantification, despite longstanding safety-science recommendations emphasizing traceability and reliability under uncertainty as articulated by Reason (1997) and later extended to socio-technical safety contexts by Leveson (2011). Climate RADAR extends this research strand by embedding Bayesian posteriors and calibrated uncertainty intervals so that risk scores can directly govern escalation thresholds and decision triggers, aligning with recent calls for explicit reliability guarantees in high-stakes decision systems (Amodei et al., 2016; Hendrycks et al., 2021).

### 7.3 Machine Learning for Hazard Forecasting

Machine learning has seen extensive application in rainfall–runoff prediction, heatwave intensity forecasting, and hydrological signal modeling, with sequential learning techniques offering notable improvements in representational capacity. Work on interpretable model behavior in complex temporal settings (Miller, 2019) and scalable multimodal forecasting (Bommasani et al., 2022) demonstrates that modern architectures can achieve strong predictive accuracy; however, the literature has rarely examined whether predictive improvements compel more timely or equitable protective actions. This gap has been repeatedly noted in responsible-AI discourse where the utility of predictive systems is framed not by error minimization but by behavioral impact and intervention success (Floridi & Cowls, 2021). Climate RADAR directly operationalizes this critique by linking uncertainty distributions to action-execution metrics—evaluating performance not in terms of RMSE or skill score alone but by reduction in response delays and improved willingness to act.

### 7.4 Generative AI, Guardrails, and Safe Deployment

LLMs have introduced multilingual reasoning, contextual judgment, and situational flexibility into high-risk decision environments, yet their benefits inherently coexist with hallucination, policy-misalignment, representational fairness issues, and unverifiable internal logic. Governance-driven safety approaches have traditionally emphasized model-level value alignment and uncertainty disclosure (Amodei et al., 2016; Hendrycks et al., 2021), while sociotechnical oversight frameworks emphasize transparency, reviewability, and ethical accountability as foundational components for trust establishment (Winfield, 2018; Raji & Buolamwini, 2020). Despite this, disaster-resilience contexts remain largely unexplored, particularly where time constraints, vulnerable user groups, and jurisdictional responsibilities intensify the risk of model failure. Climate RADAR advances prior guardrail approaches by grounding model outputs in a calibrated risk index, tracking decision provenance, and embedding escalation rules that prevent autonomous execution under uncertainty levels that exceed safety tolerances.

## 7.5 Safety Science and Regulatory Frameworks

Safety science provides a systematic foundation for operational accountability, arguing that failures in complex systems arise not from single-point mistakes but from compound breakdowns in monitoring, interpretation, and handoff mechanisms. Foundational models of systemic failure articulate this view clearly (Reason, 1997), later expanded into safety-critical engineering practice through formally audited governance logic (Leveson, 2011). In parallel, disaster-risk governance underwent codification through the Sendai Framework, which foregrounds people-centered risk reduction, action-oriented preparedness, and equitable accessibility (United Nations Office for Disaster Risk Reduction, 2015). More recently, the EU AI Act formalized disaster management decision systems as "high-risk AI," demanding transparent versioning, mandated human override, fairness accountability, and traceable evidence bundles (European Union, 2024). Climate RADAR operationalizes these requirements by attaching explicit evidence bundles to each generated recommendation, aligning system behavior with escalating human-control thresholds under safety budgets, and logging all critical decision states for audit recovery.

# 8 Future Work

While the proposed Climate RADAR framework demonstrates strong results, several avenues remain open for advancing both research and practice.

## 8.1 Toward Self-Healing Resilience

Future iterations will focus on advancing Climate RADAR toward selectively autonomous capabilities that support self-healing resilience mechanisms. Rather than full automation, the goal is to enable conditional autonomy in well-bounded scenarios—for example, issuing localized pre-alerts or pre-staging resources when uncertainty and harm potential fall within safe operational envelopes. The central research challenge is to formalize appropriate decision boundaries and escalation triggers so that operators remain clearly in control, while ensuring that safety budgets, override policies, and audit trails remain transparent, traceable, and governed under accountable human supervision (Winfield, 2023; Amodei et al., 2016; Hendrycks et al., 2021; Leveson, 2011; Reason, 1997).

## 8.2 Scaling to Multi-Hazard, Multi-City Deployments

Our current evaluation focused on flash flood and heatwave scenarios within a single municipal jurisdiction, which allowed controlled testing of risk estimation, user interaction patterns, and fairness outcomes under operational constraints. Future work will expand to multi-hazard contexts (e.g., typhoons, wildfires, earthquakes) across multiple cities and regions. Such deployments will allow us to evaluate robustness under heterogeneous infrastructures, diverse socio-demographic populations, and varying governance models Basher (2006); Birkmann et al. (2013); Cutter et al. (2003).

## 8.3 Fairness-Aware Personalization

Although subgroup fairness audits highlighted disparities, we did not yet implement algorithmic debiasing methods. Future work will incorporate fairness-aware optimization techniques, such as subgroup-specific threshold calibration, reweighting, and adversarial debiasing Jobin et al. (2019);

Weidinger et al. ([2022](#)); Raji and Buolamwini ([2020](#)). These approaches will be validated using established fairness metrics (e.g., equal opportunity difference, demographic parity gap) to ensure that vulnerable populations benefit equitably from the system Floridi and Cowls ([2021](#)).

### 8.4 Longitudinal Studies of Trust and Human–AI Teaming

Our evaluation captured short-term trust dynamics. Future research will conduct longitudinal studies to investigate sustained trust, potential fatigue effects, and user adaptation under repeated hazard events. These studies will also examine the evolving role of human–AI teaming, particularly how municipal staff calibrate reliance on automated recommendations over time Winfield ([2018](#)); Floridi and Cowls ([2021](#)); Bommasani et al. ([2022](#)).

### 8.5 Integration with Policy and Regulatory Frameworks

As policymakers enact stricter governance for high-risk AI systems, we will extend Climate RADAR's compliance features. This includes explicit mapping of system safeguards to the Sendai Framework, EU AI Act requirements, and emerging national AI policies United Nations Office for Disaster Risk Reduction ([2015](#)); European Union ([2024](#)). We also plan to collaborate with regulatory bodies to establish benchmarks for accountability, reproducibility, and fairness in disaster resilience platforms Jobin et al. ([2019](#)); Weidinger et al. ([2022](#)).

## 9 Conclusions

This study introduced Climate RADAR, a reliability layer that connects predictive risk estimation to measurable protective actions in disaster settings. Through simulations, user studies, and municipal deployment, we showed that the system increases action execution rates, reduces response time, and improves usability and perceived trust among participants. Fairness assessments further revealed that targeted personalization can narrow behavioral gaps for vulnerable subgroups.

Beyond these empirical findings, the study highlights design principles for decision-support systems operating under uncertainty. Specifically, linking calibrated risk estimates to auditable recommendations, human-override rules, and safety constraints enabled more accountable and equitable decision pathways. These insights contribute to ongoing efforts to design disaster-management infrastructures that emphasize not only situational awareness but reliable action execution. Future extensions will investigate multi-hazard deployments, long-term user adaptation, and expanded fairness interventions across diverse population groups.

**Acknowledgements**
The authors thank municipal partners and study participants for their invaluable contributions to the pilot deployment and evaluation of Climate RADAR. This work was supported by the Digital Social Innovation Service Contest from the Ministry of Science and ICT and the National Information Society Agency (NIA). The funding body had no role in the design of the study, data collection, analysis, interpretation, or in the writing of the manuscript.